\documentclass[runningheads]{llncs}
\usepackage{graphicx}
\usepackage{algorithm}
\usepackage{algorithmicx}
\usepackage{algpseudocode}
\usepackage[table]{xcolor}
\usepackage{amsmath}
\usepackage{changepage}
\usepackage{subfigure}
\usepackage[utf8]{inputenc}
\usepackage{multirow}
\usepackage{color}

\begin{document}
\title{Imputation techniques on missing values for breast cancer treatment and fertility data}
%
\titlerunning{\textit{Imputation techniques for missing values}}
%
\author{Xuetong Wu\inst{1} \and
Hadi Akbarzadeh Khorshidi \inst{1} \and
Uwe Aickelin\inst{1} \and
Zobaida Edib\inst{1} \and
Michelle Peate\inst{1}}
\authorrunning{X. Wu et al.}
%
\institute{University of Melbourne, Parkville, Victoria 3010 \\
\email\{xuetongw1,zedib\}@student.unimelb.edu.au\\
\email\{hadi.khorshidi,uwe.aickelin,mpeate\}@unimelb.edu.au}
\maketitle              

\begin{abstract}
Clinical decision support using data mining techniques offers more intelligent ways to reduce decision errors in the last few years. However, clinical datasets often suffer from high missingness, which adversely impacts the quality of modelling if handled improperly. Imputing missing values provides an opportunity to resolve the issue. Conventional imputation methods adopt simple statistical methods, such as mean imputation or  discarding missing cases, which have many limitations and thus degrade the performance of learning. This study examines a series of machine learning based imputation methods and suggests an efficient approach for preparing a good quality breast cancer dataset, to find the relationship between breast cancer treatment and chemotherapy-related amenorrhoea, where the performance is evaluated by the accuracy of the prediction. To this end, the reliability and robustness of six well-known imputation methods are evaluated. Our results show that imputation leads to a significant boost in the classification performance compared to the model prediction based on list-wise deletion. Furthermore, the results reveal that most methods gain strong robustness and discriminant power even when the dataset experiences high missing rates ($>50\%$).

\end{abstract}
\section{Introduction}
Clinical data with substantially missing information presents significant challenges for pattern classification and decision making. Machine learning and statistical analysis based clinical decision support systems associate a patient's health status with the prediction for a disease or medical outcomes of interest, such as in-hospital mortality \cite{lee2012transfer}, breast cancer \cite{jerez2010missing} and diabetes \cite{purwar2015hybrid}. Data mining has been widely recognised as a crucial approach for many clinical prediction rules. In practice, clinical datasets are sometime incomplete, usually attributed to manual collection, erroneous measurements and equipment failures. The missing values dramatically degrade the performance if handled improperly. When the missing rate exceeds $15\%$, missing values should be carefully treated with special consideration \cite{Acuna2004missing}. The simplest solution is the case deletion strategy, which discards all missing cases and works only when a few missing values exist. Another solution is substituting missing entries with the mean or mode values of a specific feature, which reduces the variability of the dataset and ignores the covariance among features \cite{schafer1997analysis}. Such techniques have many limitations and may not always benefit model construction. Our motivation for innovative imputation is to develop efficient and beneficial algorithms to improve the classification performance. Many studies have demonstrated that machine learning based techniques are effective and useful in managing small to large missingness and data scales \cite{lin2019missing}.

Over the last decade, there have been tremendous developments in clinical data analysis. Breast Cancer (BC) is the most common type of cancer in women  \cite{lee2009chemotherapy,Liem2015chemo}. Among those who are of reproductive age ($<$35 years) at diagnosis, $15\%$ have not yet been pregnant or started a family \cite{ruddy2014breast}. According to \cite{Ives2007pregnancy}, the 10-year survival rate of early BC is almost $85\%$ and pregnancy does not negatively impact prognosis. 
However, one of the main side effects of cancer treatment is amenorrhoea (permanent cessation of menstruation), that can affect up to 98$\%$ of women in the reproductive age, and around 76$\%$ of survivors wish to conceive in the future \cite{peate2017fertility}. Therefore, infertility risk prediction after cancer treatment becomes a priority for young BC patients who wish to conceive after cancer. In this study, we are preparing datasets to describe factors related to fertility and breast cancer treatments to determine likely post-treatment amenorrhoea. The datasets are collected from six international institutions and were originally archived in different formats, which are not fully aligned e.g. some essential features in one subset may be fully or partially absent in another. As a consequence, substantial missing values are introduced.

We aim to evaluate some well-known imputation methods, i.e., mean/mode imputation, random imputation, multiple imputation using chained equation (MICE) \cite{buuren2010mice}, k-nearest neighbour (KNN) \cite{batista2002study}, random forest (RF) \cite{stekhoven2012miss} and expectation maximisation (EM) \cite{Moon1996em} imputation of missing values and find the potential relationship between cancer treatments and post-treatment amenorrhoea by constructing multiple classifiers. Use of the datasets are authorised by the FoRECAsT consortium of Psychosocial Health and Well-being Research (emPoWeR) Unit, University of Melbourne \cite{FoRECAsT}. The quality of the imputation will be measured by the prediction accuracy of the post-treatment amenorrhoea status. To this end, we undertake extensive experimental comparisons and simulations with some popular imputation algorithms. The main contributions of this paper are as follows:

1. The work explores the impact of some notable imputation techniques using statistical and machine learning methods, and further evaluates the performance regarding the classification tasks on prediction of amenorrhoea after 12 months of breast cancer treatments. 

2. We examine whether the imputation across different datasets achieves significant improvements, even if the data has a large amount of missing values.

The paper is structured as follows. Related work of imputation techniques and chemotherapy-related amenorrhoea are illustrated in section 2. Section 3 reviews the clinical data and includes descriptions of experimental methodologies. Section 4 presents and discusses the results. Section 5 concludes the paper.

\section{Related Work}
Chemotherapy-related amenorrhoea (CRA) can be caused by breast cancer treatment, the maintaining of fertility options should be assessed before the treatment regularly and young women should be informed of the possibility of amenorrhoea or recovery of menstruation and contraceptive choices \cite{peate2011}.
Lee \cite{lee2009chemotherapy} reported that the incidence of CRA hinges on age at diagnosis and adjuvant endocrine therapy, for those who are older than 40 years, CRA is more likely to occur and be permanent, especially after adjuvant endocrine therapy. Also Liem \cite{Liem2015chemo} pointed out the age at diagnosis is the main factor associated with chemotherapy-related infertility. Apart from the age, post-cancer fertility will also depend on personal factors, Peate \cite{peate2011} found out that low knowledge can reduce the quality of decision making. To conclude, prediction of chemotherapy-related infertility involves consideration of complex factors such as age, lifestyle factors, knowledge, previous pregnancies, ovulation, history of previous medical and gynaecological diseases \cite{johnson2006ovarian}. Decision support is critical in ensuring patients can make informed decisions about fertility preservation in a timely manner, but in practice women are making this decision without knowing their infertility risk, which has the potential for adverse effects. The key challenge with fertility prediction is that the data usually contains substantial missing elements which adversely impact the prediction results. Imputation methods can help to accommodate this issue.

\subsection{Missing values}
\noindent Missing values are common in datasets and this has serious drawbacks for data analysis. The reasons for missing data may vary, some information cannot be obtained immediately, data might be lost due to unpredictable factors, or the cost for accessing the data is unaffordably high. There are four main types of approaches for dealing with missing data, these include deleting the incomplete data and only use complete data portions, treating missing values as a new category where standard routines can be applied, using statistical based procedures, e.g. mean imputation and EM algorithms, and adopting machine learning methods, such as KNN, decision trees, and logic regression method.

Different types of missing data are defined by Little and Rubin \cite{little2019statistical}, who categories missing data into three types, which are \textit{missing completely at random}(MCAR), \textit{missing at random}(MAR) and \textit{missing not at random}(MNAR).

MCAR cases occur when the probability that an element is missing is independent of the variable itself or any other related influences, simple examples of MCAR include accidental data lost, occasional omission collection of questionnaire, and manual recording errors in medical data. MAR is the case such that the missing value is independent of the missing attribute itself but can be predicted from the observed responses. A typical case is that young BC patients have more missing data in terms of fertility and productivity, compared with older patients, by leveraging the observed age information. MNAR situation occurs when the missingness is related to the missing feature itself and missing data cannot be predicted only from the observed and missing entries from the database. For example, BC patients will be more inclined to conceal private information unrelated to the cancer such as education and salary levels, which are unlikely to be foreseen. Handing this category of missing data is problematic and there are no generalised methods that can resolve this issues properly. 

In our case, the MNAR type is rare amongst the data types \cite{Moniek2013missing}, so we only consider that the missing values under MCAR or MAR assumptions unless a feature is totally missing.

\subsection{Imputation Techniques}
\subsubsection{Imputation with statistical analysis}
When missing data are MCAR or MAR, they are termed 'ignorable' or 'learnable', which implies that researchers can impute data with certain procedures, by statistical analysis or machine learning approaches. Some popular and well-known statistical imputation techniques will be presented in this section.
The easiest way of filling missing value is imputing the average value of the observed data, known as 'mean imputation'. The method is elementary, but the drawbacks are obvious, e.g., it fails to deal with large amount of missing values, distorts the distribution of the dataset and totally ignores the covariance between different attributes as the expectation of the attribute $E[x_i]$ does not change in this case \cite{schafer1997analysis}. Another imputation method, assigning a random selection of observed values for missing items \cite{Kalton1984random}, is also employed when missing features are numerical. However this method neglects the latent potential relationship among present features.
Another way to utilise the knowledge of whole dataset is the model-based approach. Expectation maximum (EM), is introduced to deal with missing data by fitting models to the incomplete data capitalising on the knowledge from complete sets \cite{Nelwamondo2007miss}. If the model is correct for the complete sample, the maximum likelihood estimation of the unknown parameters can be made by observing the marginal distribution of the data \cite{little2019statistical}. Expectation maximisation is suitable and outperforms mean substitution and list-wise deletion in cases where there is little or no interdependency between the input variables.
The methods described above are single imputation and the statistical uncertainty of the missing values is not reflected. The multiple imputation (MI) procedure provides a solution to this issue \cite{rubin2004multiple}, MI replaces all missing values with a set of conceivable attributes that can present the uncertainty of the plausible values which are generated by regression models. All missing data is filled in $M$ times ($M > 20$) to generate $M$ complete data sets. The $M$ complete datasets are then analysed for certain standard tasks such as regression, classification and clustering. The results are pooled and averaged to produce a single estimate. MICE imputation is particularly flexible in a broad range of frameworks as it invents varied complete datasets and takes the uncertainty into account to yield accurate deviations. But as the prediction models are constructed successively, the computational cost is relatively high.

\subsubsection{Imputation with machine learning methods}
Machine learning achieves great success in many fields and the flexibility allows us to capture the high-order interactions in the data \cite{jerez2010missing} and thus impute missing attributes. This section reviews several imputation routines characterised by machine learning concepts.

KNN approach is a type of hot deck supervised learning method, providing a path to find the most similar cases for the given instances, in which KNN is a useful algorithm that matches a case with its closest $k$ neighbours in a multi-dimensional space. In missing data imputation, KNN aims to find the nearest neighbours to minimize the heterogeneous euclidean-overlap metric distance \cite{wilson1997improved} between two samples, then missing items are further substituted with the values from $k$ complete cases. The advantage is that the method is suitable for large amount of missing data, but the disadvantage is the high imputation cost as it will compare all dataset and find the most similar cases.
Moreover, Stekhoven \cite{stekhoven2012miss} proposed a model-based iterative imputation method based on random forest. The random forest is generated by decision trees from sampled subset of datasets, the proximity matrix from the random forest is learned and updated to approximate the missing values while a set of fitting models are constructed. The random forest imputation can deal well with non-parametric data with mixed types.
In this study, we will compare six common imputation techniques, which are mean imputation, random imputation, multiple imputation using chain equations, expectation imputation, KNN imputation and random forest imputation, where the classification accuracy is measured for different datasets, compared with the outcome resulting in raw data by list-wise deletion. Finally we examine whether the imputation works across different datasets.

\section{Experiments}
\subsection{Experimental setup}
\subsubsection{Data description}
\noindent The FoRECAsT dataset is split into six sub-datasets, and the basic information is summarised in Table \ref{obrate}: it contains 1565 records and 87 features. The six subsets are collected from different collaborators all over the world and combined in regulated formats, e.g., all age data are grouped into a particular feature. Features of interest are not fully aligned, e.g., some features are  observed in one sub-set  but totally absent in another, which will introduce large missingness in the whole dataset.

Among all entries, 37443 are observed and 98712 are missing, presenting 72.5$\%$ missing values in the whole dataset. The main features for mining include personal health status and some cancer-oriented features such as age category, smoking status, alcohol intake, body mass index (BMI) classes, and pregnancy-related status. The according outcome label of interest is the amenorrhoea status after the cancer treatment for 12 months, which are binary indicators, where 0 stands for negative status and 1 stands for positive. The outcome label is abbreviated as \textit{'Amen\_ST12'} and it is totally complete (100$\%$ observation rate).
\begin{table}[H]
\centering
\caption{Data description for FoRECAsT dataset, the dataset is split into six subsets regarding the sources. Observed feature here implies that the feature has at least one observation within the dataset, and missingness hinges on the observed features.}\label{obrate}
\scalebox{0.9}{
\begin{tabular}{|c|c|c|c|c|c|c|}
\hline
Data track & Instances &Observed Features & Categorical & Numerical & Missingness & Label \\
\hline
Track 1  & 725 & 19 & 19 & 0 & 8.6$\%$ & \multirow{6}*{\textit{'Amen\_ST12'}}\\
Track 2 & 280 & 36 & 36 & 0 & 9.1$\%$ & \\
Track 3 & 209 & 34 & 29 & 5 & 10.5$\%$ &\\
Track 4 & 154 & 20 & 20 & 0 & 22.2$\%$ & \\
Track 5 & 101 & 42 & 40 & 2 & 23.6$\%$ &\\
Track 6 & 96 & 47 & 43 & 4 & 18.3$\%$ &\\
\hline
Total & 1565 & 87 & 76 & 11 & 72.5$\%$&\\
\hline
\end{tabular}}
\end{table}

\subsection{Method}
The proposed method consists of two phases, imputation and prediction process. In the imputation procedures, we firstly test six common imputation methods on single data tracks separately to work out whether prediction can take advantage of filling missing items, and extended imputation experiments are conducted on the whole dataset by applying the different present features, namely 'cross imputation'. The purpose of the experiments is to investigate the robustness of the imputation techniques and potentially find better correlation between cancer treatment and fertility, by intelligently utilising missing values across different datasets, rather than individual ones. Regarding the prediction procedures, a collection of common supervised learning classifiers are formed and the results of 5-fold cross-validation predication accuracy are discussed. 
\subsubsection{Imputation and Classification methods}
To investigate the effectiveness of the imputation methods on infertility classification tasks, six notable approaches are adopted, which include mean imputation, random imputation, MICE, EM, KNN and RF imputation. All modules are implemented in Python 3.7 \cite{van1995python} in the Mac OS 10.14.3  operating system.

The datasets are a mixture of types, including quantitative and qualitative data where RF imputation generally fits well. Specially for mean imputation, absent numerical data are substituted with means while categorical cases are replaced with their mode.  Random imputation and KNN imputation will fill missing values with possible selections from observed cases. Furthermore, two model-based algorithms, EM and MICE will learn series of classifiers and regressors for categorical data and numerical, respectively. 

From the perspectives of prediction procedures, six common supervised learning classifiers are constructed to test the effectiveness of imputation. 
The classification model include support vector machine (SVM), decision tree (DT), multilayer perceptron (MLP), random forest (RF), logistic regression (LR), Gaussian Naive Bayesian (GNB), and KNN algorithms. 
We undertook a series of imputation approaches on the six sub-datasets as described, and our benchmarks are the prediction outcomes from the raw data by the list-wise deletion. Extended simulations on the entire database (all instances included) using the cross imputation will also be reviewed.
\section{Result and discussion}
\subsection{Imputation on single sub-dataset}
\begin{figure}[H]
\centering
\subfigure[Track 3]{
\includegraphics[width=5cm,height = 3.8cm]{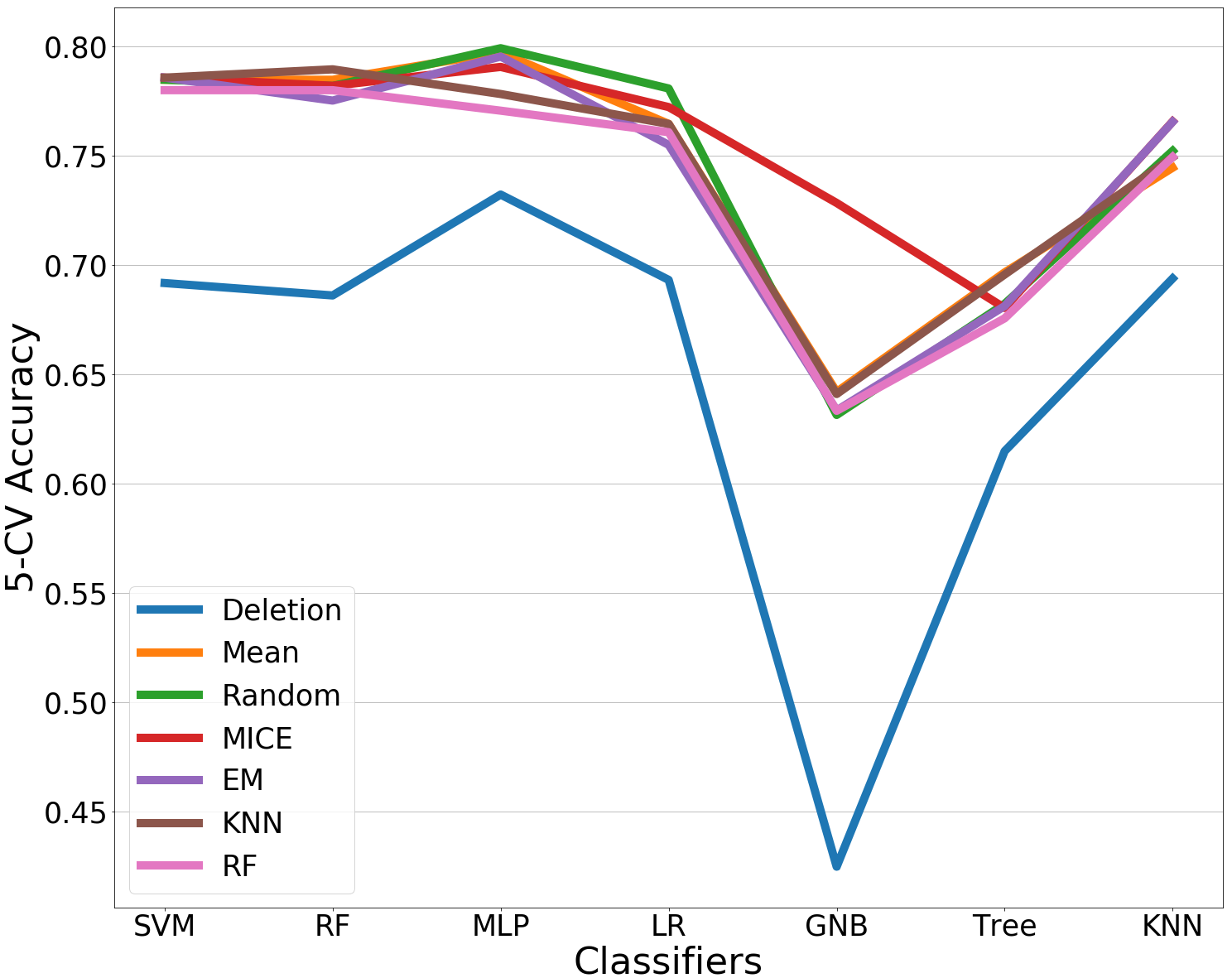}
}
\quad
\subfigure[Track 4]{
\includegraphics[width=5cm,height = 3.8cm]{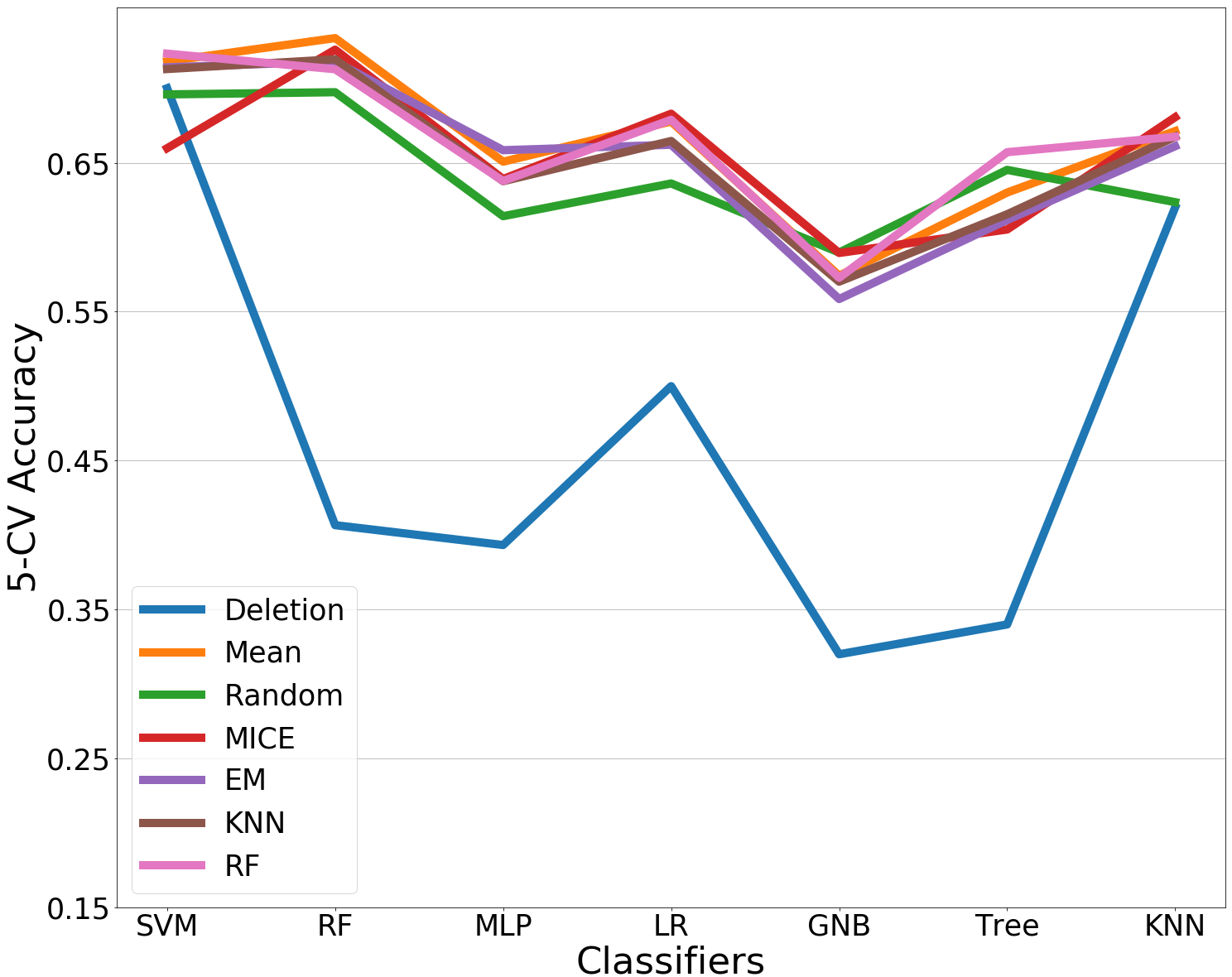}
}
\quad
\subfigure[Track 5]{
\includegraphics[width=5cm,height = 3.8cm]{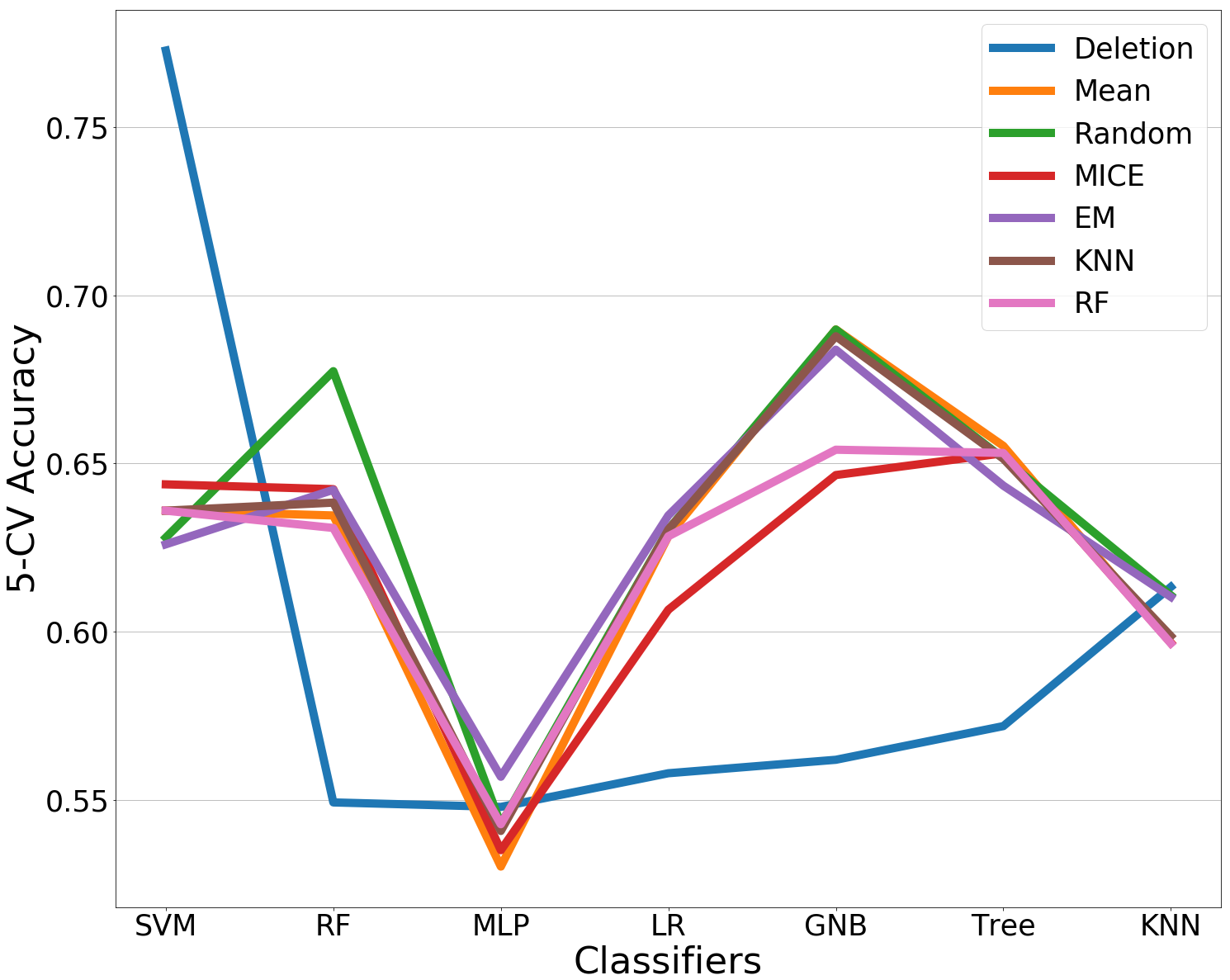}
}
\quad
\subfigure[Track 6]{
\includegraphics[width=5cm,height = 3.8cm]{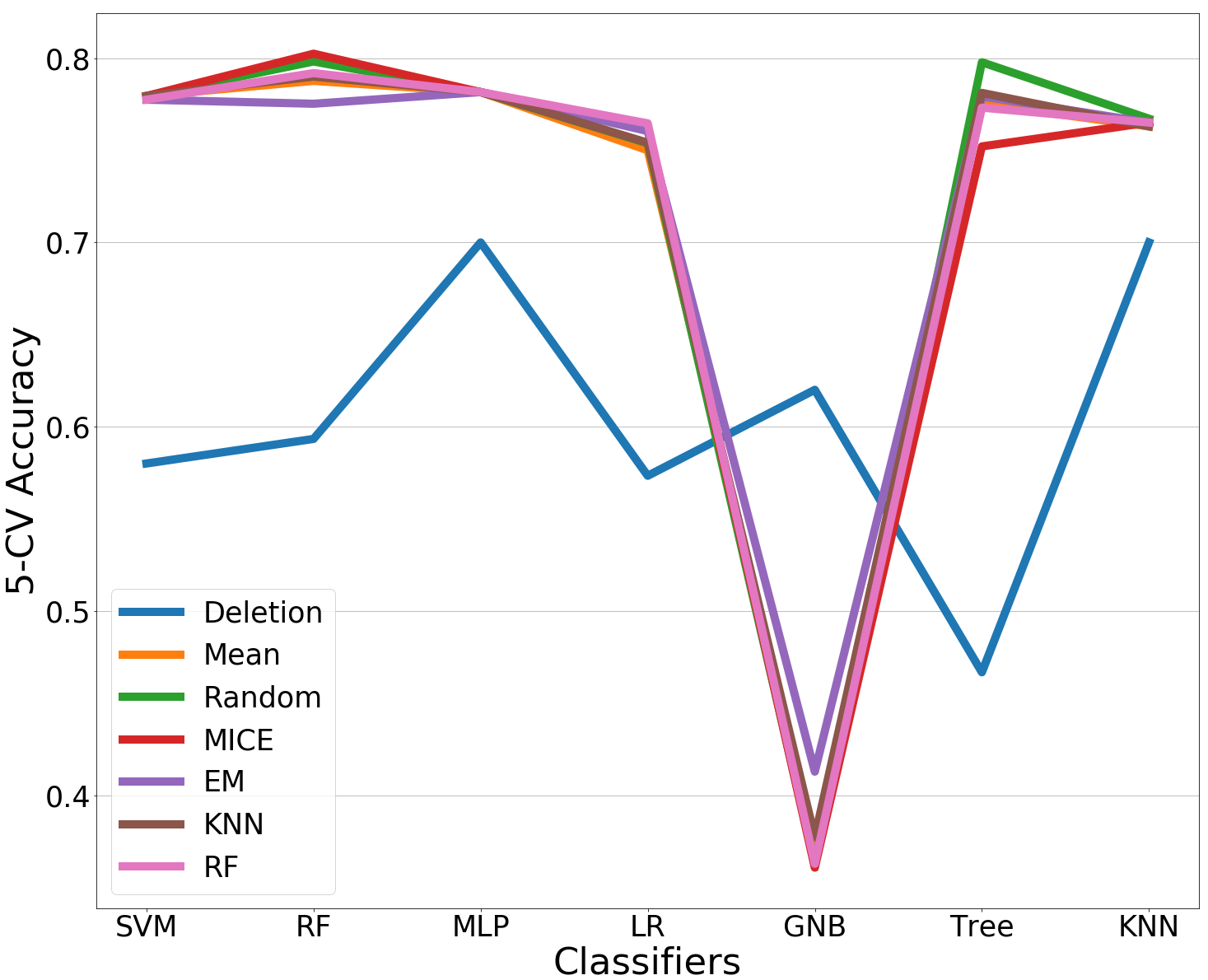}
}
\caption{Resutls of prediction (Y axis) by seven classifiers(X axis) in terms of six sub-datasets, the imputation is performed on single sub-dataset. }\label{result}
\end{figure}

Figure \ref{result} and Table \ref{resulttable} demonstrated the accuracy measured by seven classifiers and six imputation techniques, for six data tracks respectively. The imputation is implemented within the single dataset and missingness of each set is relatively low, missing values are under the assumption of MCAR or MAR. 

From the point view of classifiers, it can be seen that the SVM and RF prediction models achieve the highest accuracy, at approximately 74$\%$ in average and the number is almost 24$\%$ percentage higher than Naïve Bayesian method (50$\%$). The low accuracy of GNB is because the input space is categorical and as a result the distribution of most features is not Gaussian. 
On the other hand, by considering the imputation techniques, the results show a commonly observed pattern that most imputation methods help improve the classification performance, except a few conditions such as GNB result in track 6, MLP result in track 5. It is noticeable that even the elementary mean imputation can improve the performance to some extent and produce competitive results compared to other more complicated approaches, which emphasises the importance of filling missing values and corresponds to the results from \cite{lung2017}. 
From the average results of the classifiers, RF and MICE achieved the best performance and RF showed more robustness in prediction.  

In conclusion, when data are under MCAR or MAR cases and missingness is comparatively low, imputation is necessary as even fundamental methods can improve the prediction performance. The appropriate combination of imputation and classifier can lead to flexible and outstanding solutions for missing data.
\begin{figure}[H]
\centering
\subfigure[Track 1]{
\includegraphics[width=5cm,height = 3.8cm]{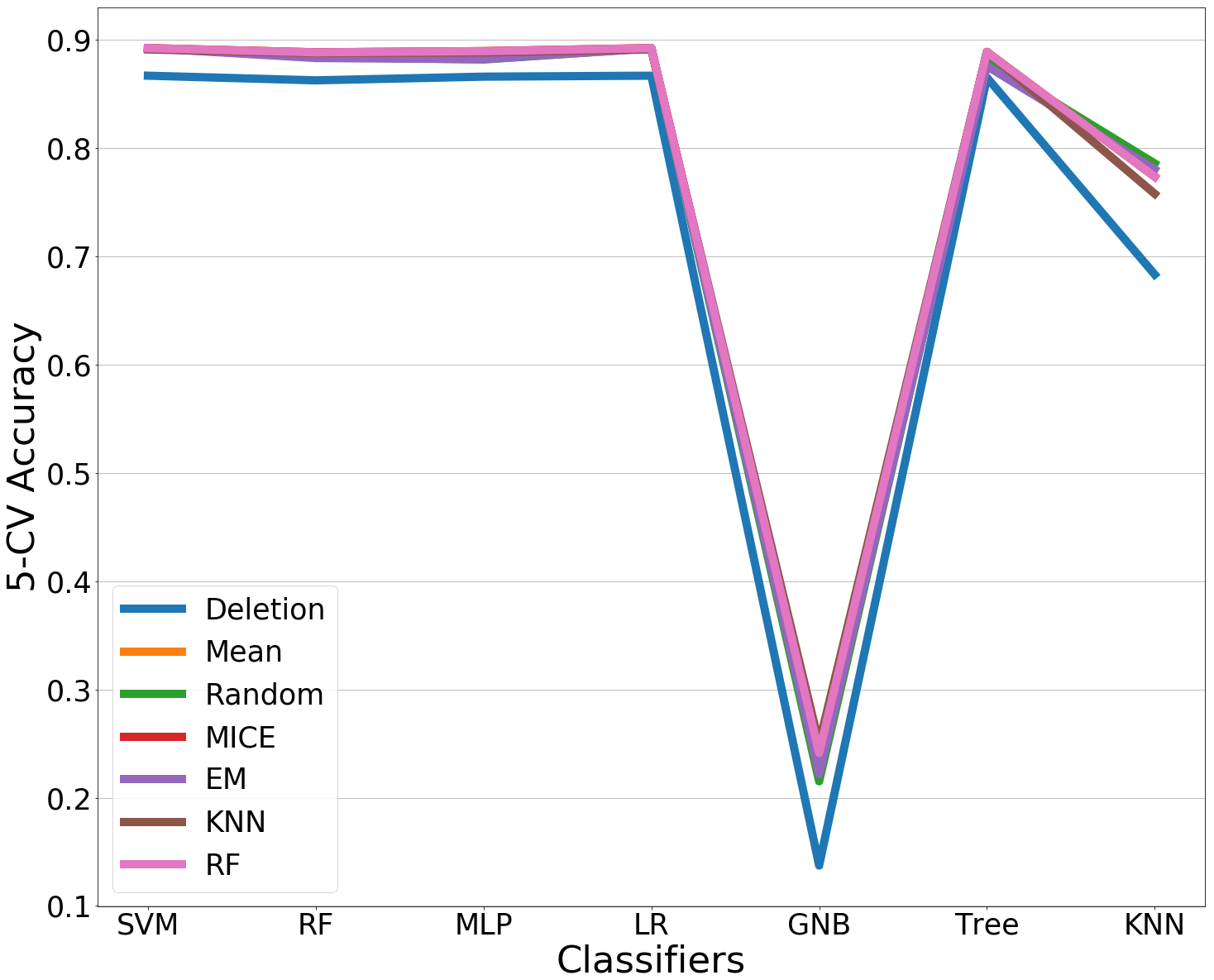}
}
\quad
\subfigure[Track 2]{
\includegraphics[width=5cm,height = 3.8cm]{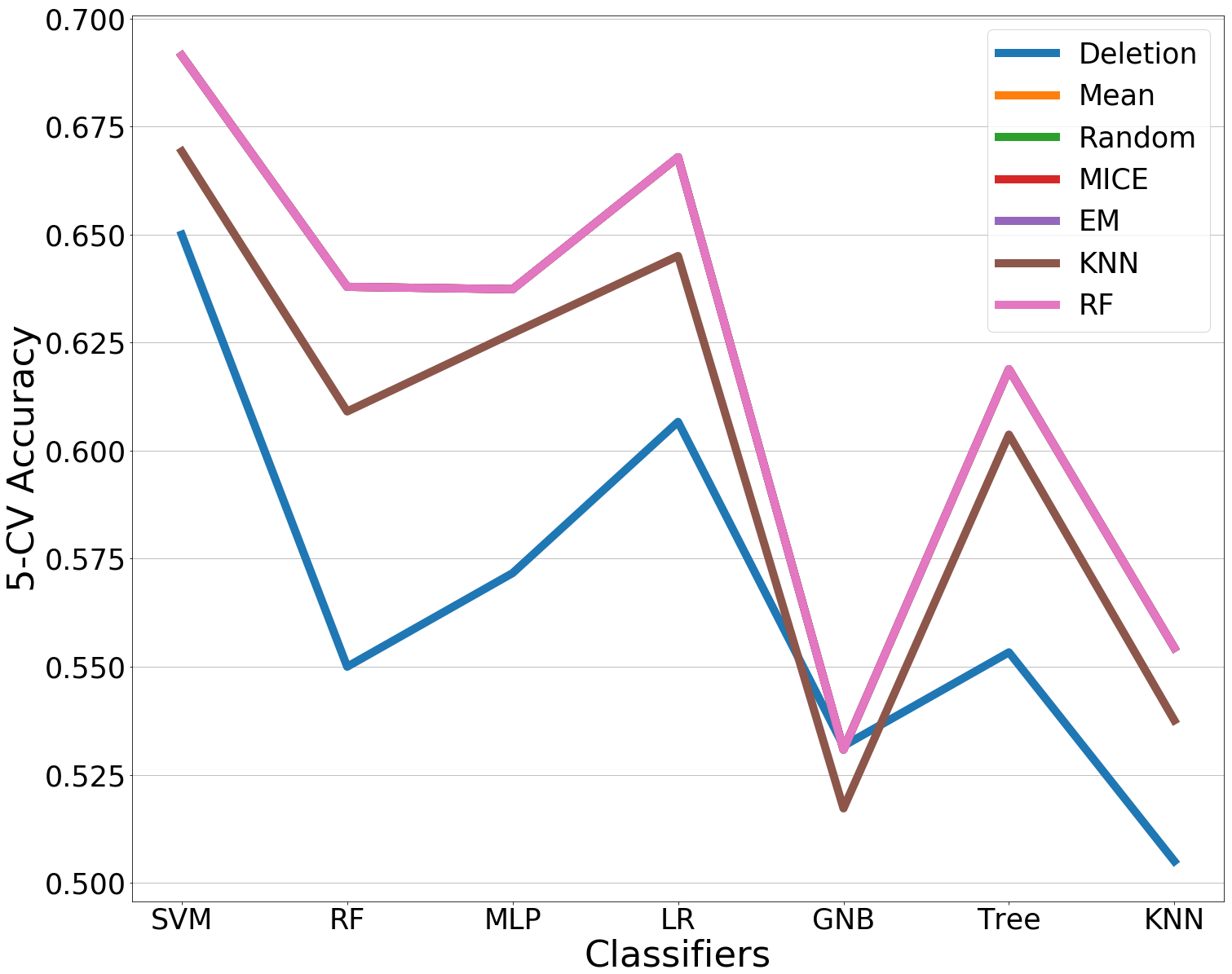}
}
\quad
\end{figure}

\begin{table}[H]
\centering
\caption{Classification accuracy and its standard deviation in different imputed datasets (upper) and classifiers (lower), the results are averaged by seven classifiers (upper) and six datasets (lower), respectively, using 5-fold cross-validation. Highest accuracy and lowest standard deviation are highlighted in bold in each row.}\label{resulttable}
\scalebox{0.77}{
\begin{tabular}{lccccccc}
\hline
\hline
\multicolumn{8}{c}{Accuracy averaged by the results from different classifiers (single dataset)} \\
\hline
Data track & Deletion & Mean & Random & MICE & EM & KNN & RF \\
\hline
Track 1 & 0.736$\pm$0.253 & \bf{0.781$\pm$0.0223} & 0.775$\pm$0.224 & 0.781$\pm$0.224 & 0.777$\pm$0.230 & 0.779$\pm$0.221 & \bf{0.781$\pm$0.223} \\
\hline
Track 2 & 0.567$\pm$0.045 & \bf{0.621$\pm$0.055} & 0.620$\pm$0.055 & 0.620$\pm$0.055 & 0.620$\pm$0.055 & 0.601$\pm$0.055 & 0.620$\pm$0.055 \\
\hline
Track 3 & 0.648$\pm$0.095 & 0.745$\pm$0.055 & 0.740$\pm$0.063 & 0.686$\pm$0.155 & \bf{0.747$\pm$0.055} & 0.743$\pm$0.054 & 0.741$\pm$0.056 \\
\hline
Track 4 & 0.469$\pm$0.134 & 0.665$\pm$0.045 & 0.656$\pm$0.055 & \bf{0.669$\pm$0.053} & 0.650$\pm$0.045 & 0.655$\pm$0.045 & 0.665$\pm$0.045 \\
\hline
Track 5 & 0.596$\pm$0.077 & 0.624$\pm$0.045 & 0.614$\pm$0.045 & 0.615$\pm$0.032 & \bf{0.627$\pm$0.045} & 0.626$\pm$0.045 & 0.616$\pm$0.033 \\
\hline
Track 6 & 0.605$\pm$0.077 & 0.716$\pm$0.141 & \bf{0.719$\pm$0.145} & 0.714$\pm$0.141 & 0.719$\pm$0.145 & 0.718$\pm$0.137 & 0.716$\pm$0.146\\
\hline
\hline
\multicolumn{8}{c}{Accuracy averaged by the results from different datasets (single dataset)} \\
\hline
Classifier  & Deletion & Mean & Random & MICE & EM & KNN & RF \\
\hline
SVM & 0.710$\pm$0.091 & 0.750$\pm$0.081 & 0.745$\pm$0.084 & 0.741$\pm$0.087 & 0.745$\pm$0.084 & 0.746$\pm$0.084 & \bf{0.751$\pm$0.081} \\
\hline
RF & 0.608$\pm$0.141 & 0.745$\pm$0.089 & 0.741$\pm$0.092 & \bf{0.747$\pm$0.090} & 0.738$\pm$0.088 & 0.739$\pm$0.095 & 0.742$\pm$0.089 \\
\hline
MLP & 0.635$\pm$0.151 & \bf{0.714$\pm$0.120} & 0.708$\pm$0.121 & 0.711$\pm$0.117 & 0.701$\pm$0.129 & 0.709$\pm$0.117 & 0.711$\pm$0.124 \\
\hline
LR & 0.633$\pm$0.120 & 0.730$\pm$0.087 & 0.727$\pm$0.091 & 0.731$\pm$0.089 & 0.726$\pm$0.088 & 0.725$\pm$0.091 & \bf{0.731$\pm$0.088} \\
\hline
GNB & 0.433$\pm$0.164 & 0.509$\pm$0.155 & 0.497$\pm$0.154 & \bf{0.519$\pm$0.169} & 0.501$\pm$0.155 & 0.507$\pm$0.151 & 0.501$\pm$0.152 \\
\hline
Tree & 0.569$\pm$0.160 & 0.711$\pm$0.095 & 0.705$\pm$0.098 & 0.710$\pm$0.095 & 0.709$\pm$0.092 & 0.706$\pm$0.100 & \bf{0.712$\pm$0.094} \\
\hline
KNN & 0.636$\pm$0.068 & 0.684$\pm$0.084 & 0.687$\pm$0.089 & \bf{0.691$\pm$0.085} & 0.683$\pm$0.093 & 0.679$\pm$0.087 & 0.687$\pm$0.086\\
\hline
\end{tabular}}
\end{table}
\subsection{Cross imputation across sub-datasets}
In this section, we use a subset of dataset to impute missing values in other subsets and we call it as \textit{cross imputation}. 
\begin{figure}[H]
\centering
\subfigure[Track 1]{
\includegraphics[width=5cm,height = 3.7cm]{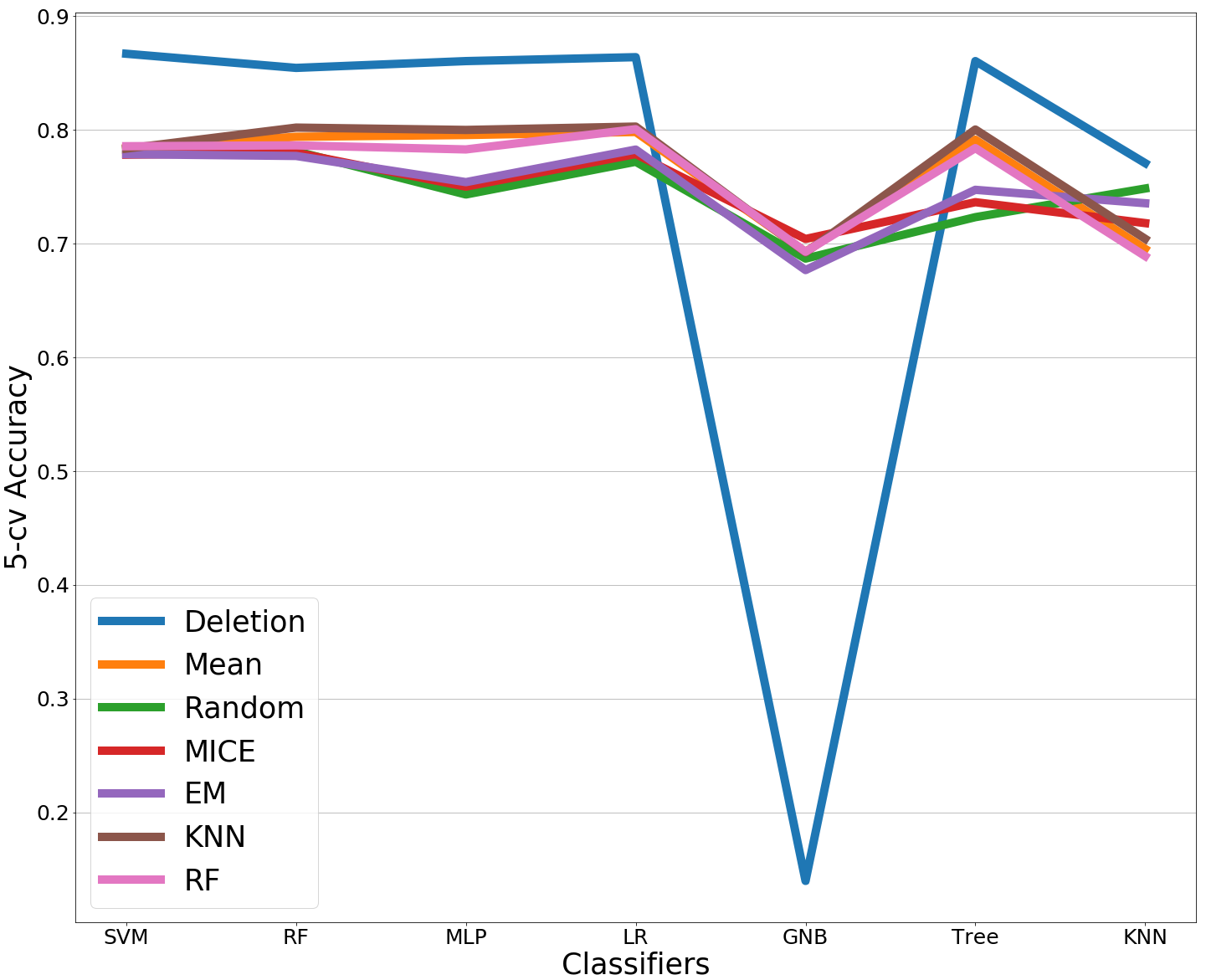}
}
\quad
\subfigure[Track 2]{
\includegraphics[width=5cm,height = 3.7cm]{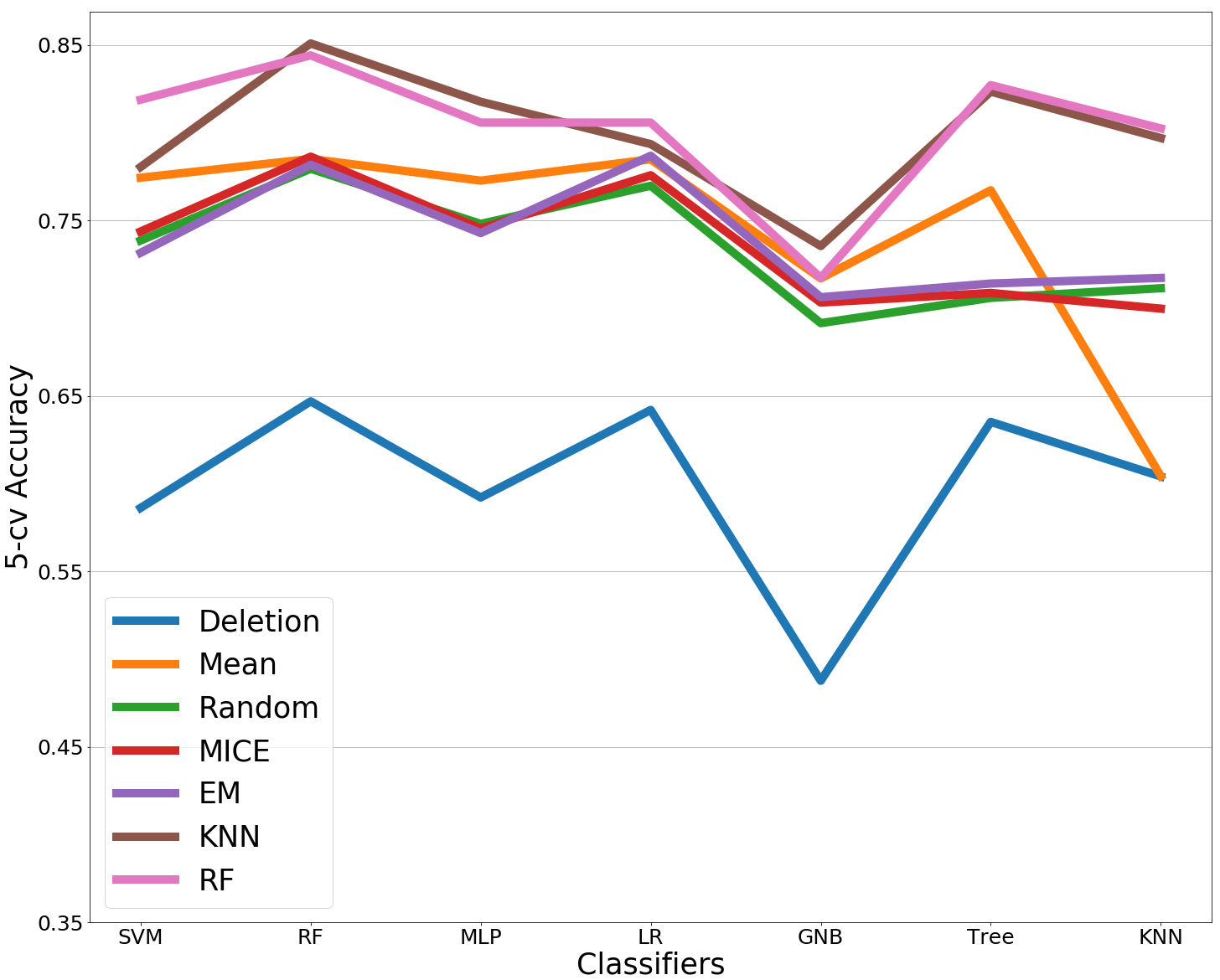}
}
\centering
\subfigure[Track 3]{
\includegraphics[width=5cm,height = 3.7cm]{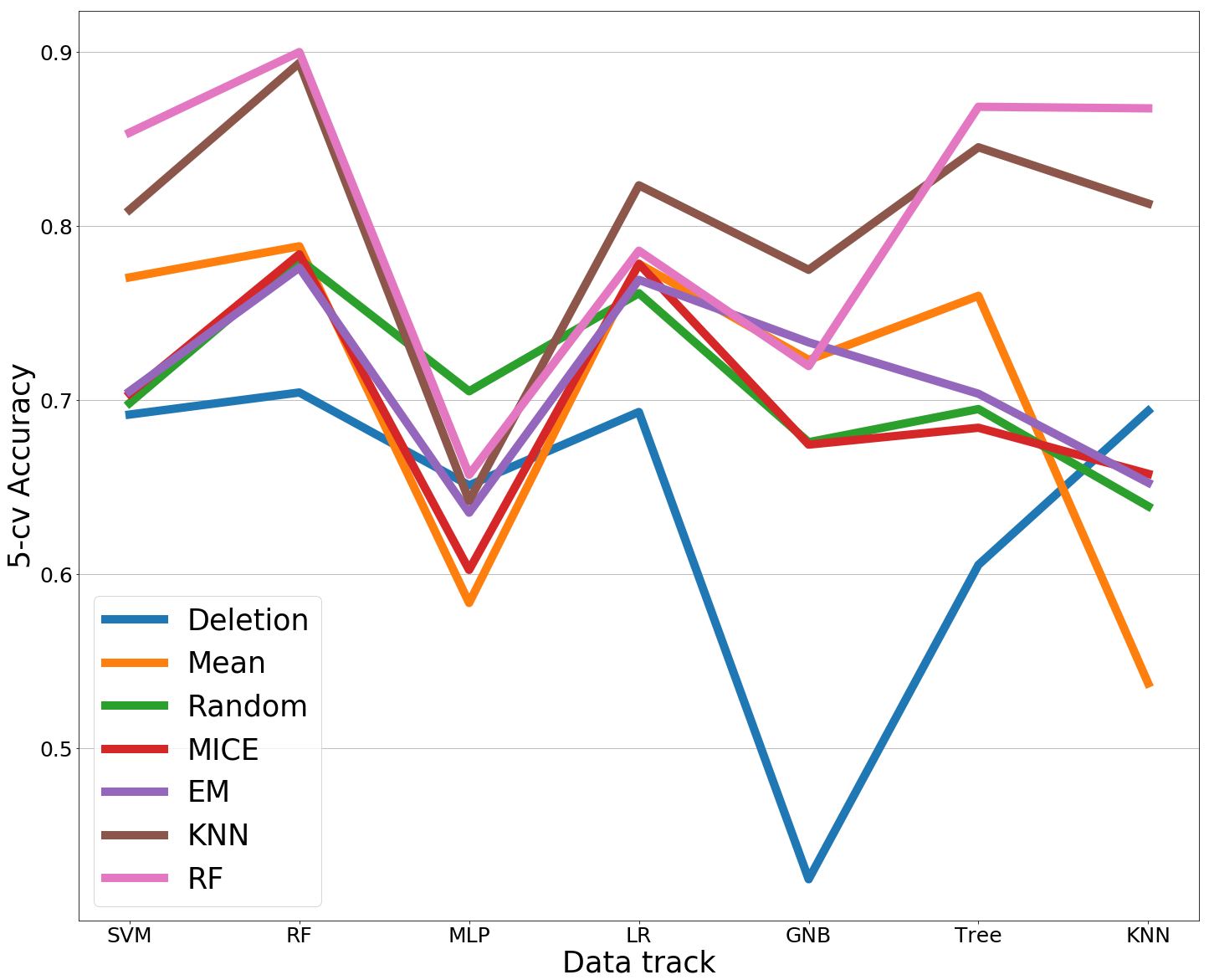}
}
\quad
\subfigure[Track 4]{
\includegraphics[width=5cm,height = 3.7cm]{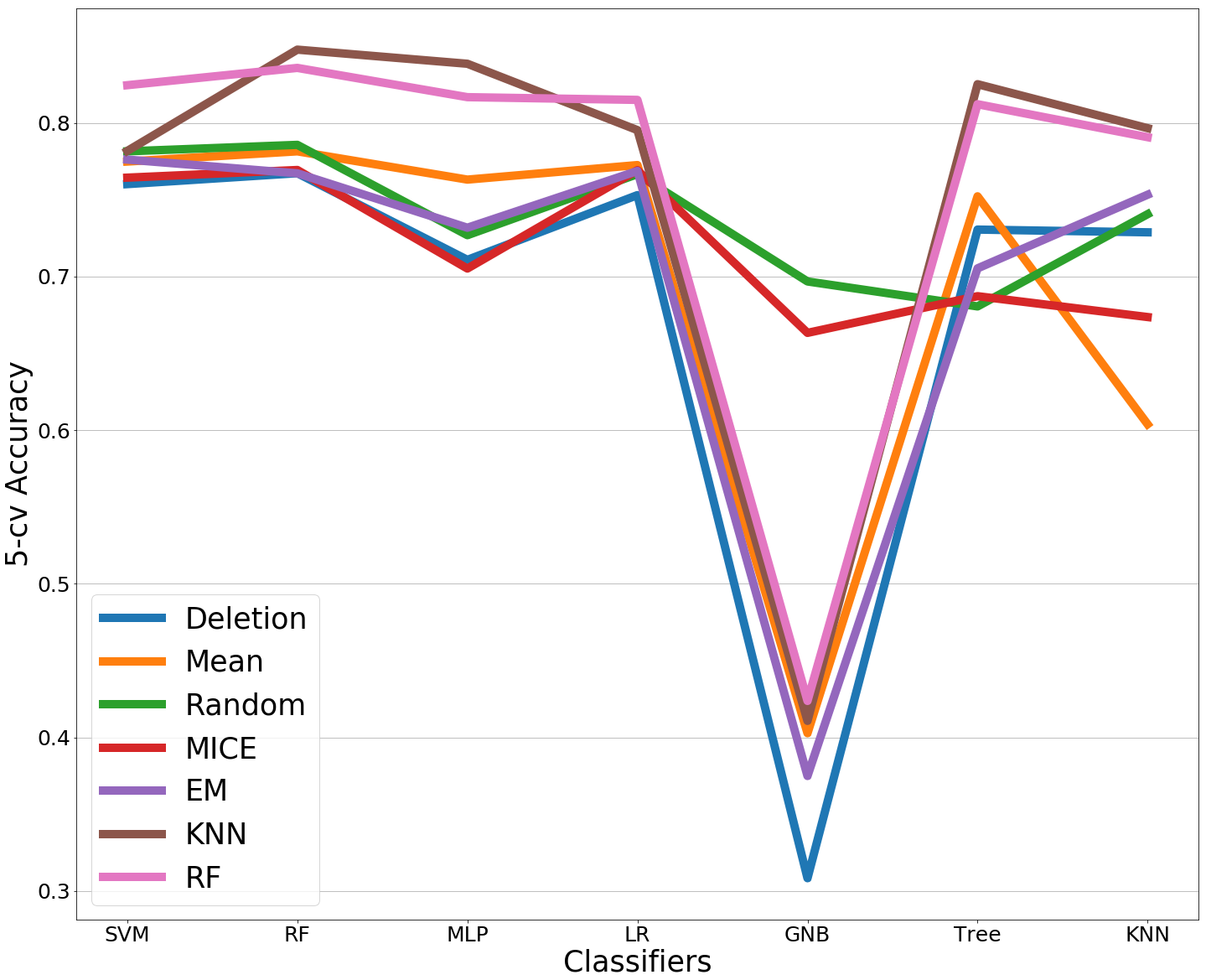}
}
\quad
\subfigure[Track 5]{
\includegraphics[width=5cm,height = 3.7cm]{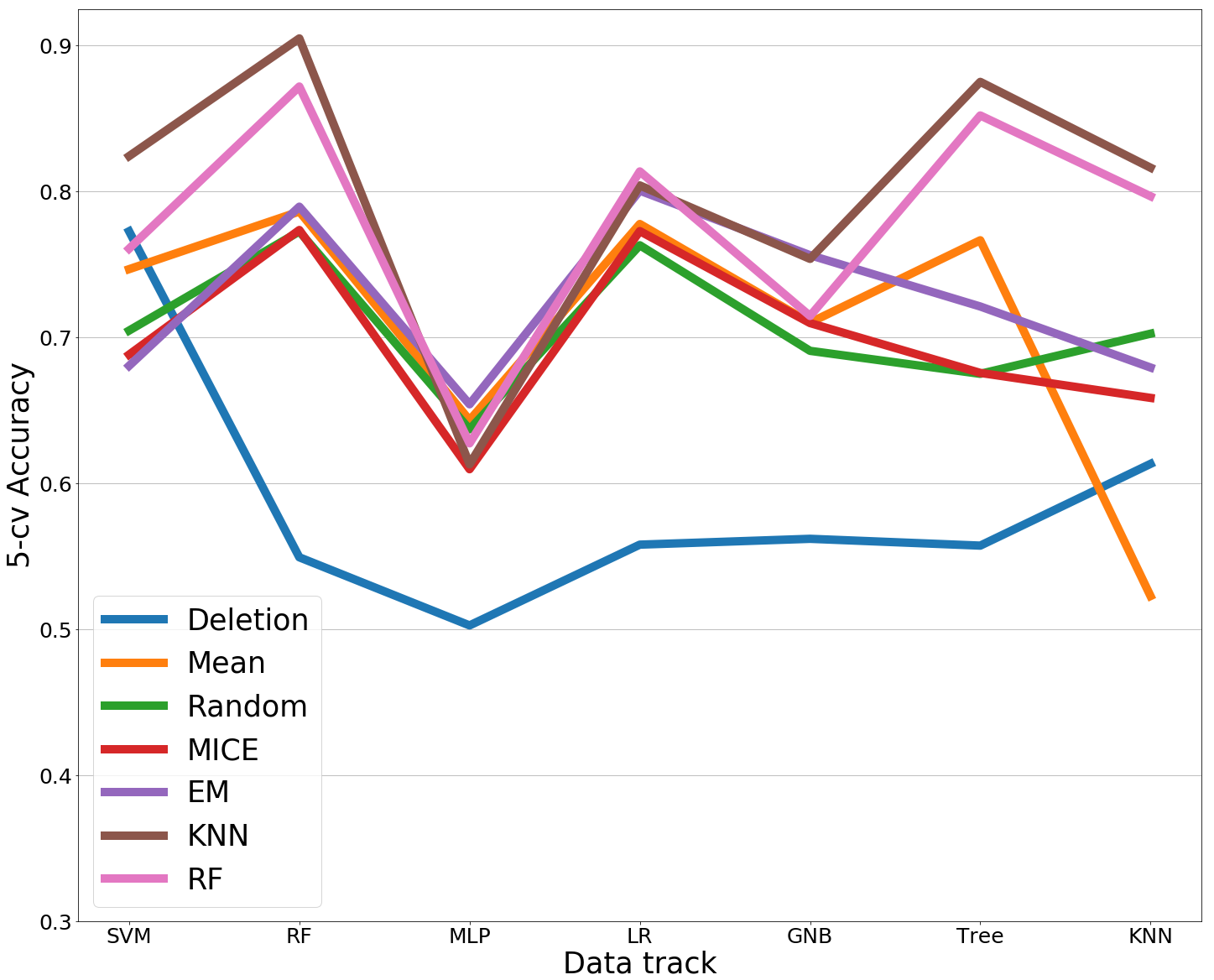}
}
\quad
\subfigure[Track 6]{
\includegraphics[width=5cm,height = 3.7cm]{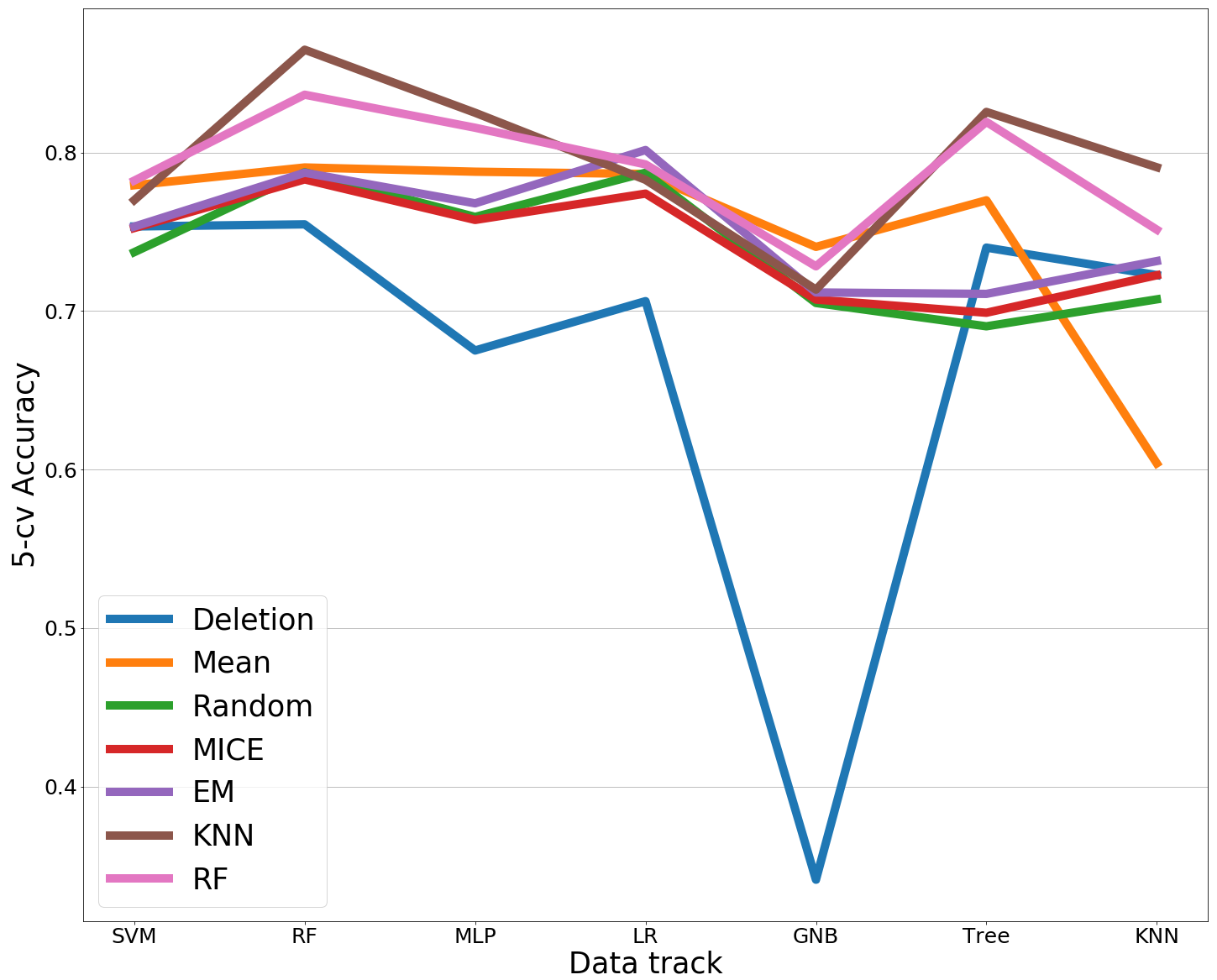}
}
\caption{Resutls of prediction(Y axis) by seven classifiers(X axis) in terms of present features from various sub-datasets, the imputation is implemented by present features in different tracks using the entire database. }\label{wholeresult}
\end{figure}For cross imputation conditions, some present features in one subset may be completely missing in other subsets, the missing information is no longer under MAR or MCAR mechanism as the researcher did not have initials in collecting them. For example, track 1 has 725 records and 19 features are observed, but for the other tracks, features are not fully overlapped with track 1 and some of which may be entirely missing, then we learn a model with present features from track 1 to impute the missing features in the rest of the sub-datasets. As a consequence, this imputation will introduce more uncertainty. Cross imputation provides a possible solution for initialising the missing values under MNAR. We examine whether the imputation techniques can still perform well in the target dataset. The results are portrayed in Figure \ref{wholeresult} and Table \ref{whole}. The differences in deletion results between single and cross imputation are compared.
\begin{table}[H]
\centering
\caption{Classification accuracy and its standard deviation in imputed whole dataset utilising cross imputation with different sub-datasets (upper) and classifiers (lower), the results are averaged by seven classifiers (upper) and six datasets (lower), respectively, using 5-fold cross-validation.}\label{whole}
\scalebox{0.73}{
\begin{tabular}{lcccccccc}
\hline
\hline
\multicolumn{9}{c}{Accuracy averaged by the results from different classifiers (whole dataset)} \\
\hline
Data track & Deletion & Mean & Random & MICE & EM & KNN & RF& Missingness \\
\hline
Track 1 & 0.745$\pm$0.249 & 0.764$\pm$0.045 & 0.748$\pm$0.032 & 0.749$\pm$0.029 & 0.750$\pm$0.034 & \bf{0.769$\pm$0.045} & 0.760$\pm$0.044 & 27.0$\%$\\
\hline
Track 2& 0.599$\pm$0.051 & 0.744$\pm$0.061 & 0.735$\pm$0.031 & 0.738$\pm$0.032 & 0.740$\pm$0.03 & 0.8$\pm$0.034 & \bf{0.803$\pm$0.038} & 46.5$\%$\\
\hline
Track 3 & 0.638$\pm$0.093 & 0.706$\pm$0.095 & 0.708$\pm$0.045 & 0.698$\pm$0.06 & 0.711$\pm$0.05 & 0.8$\pm$0.073 & \bf{0.807$\pm$0.084} & 50.2$\%$ \\
\hline
Track 4 & 0.680$\pm$0.153 & 0.693$\pm$0.132 & 0.740$\pm$0.038 & 0.719$\pm$0.044 & 0.697$\pm$0.133 & 0.757$\pm$0.143 & \bf{0.760$\pm$0.138} & $46.1\%$\\
\hline
Track 5 & 0.588$\pm$0.081 & 0.708$\pm$0.088 & 0.707$\pm$0.044 & 0.698$\pm$0.055 & 0.726$\pm$0.053 & \bf{0.799$\pm$0.088} & 0.777$\pm$0.078& $65.7\%$\\
\hline
Track 6 & 0.670$\pm$0.137 & 0.751$\pm$0.062 & 0.739$\pm$0.037 & 0.742$\pm$0.03 & 0.752$\pm$0.033 & \bf{0.796$\pm$0.045} & 0.789$\pm$0.036 & $48.2\%$\\
\hline
\hline
\multicolumn{9}{c}{Accuracy averaged by the results from different datasets (whole dataset)} \\
\hline
Classifier & Deletion & Mean & Random & MICE & EM & KNN & RF \\
\hline
SVM & 0.739$\pm$0.085 & 0.772$\pm$0.012 & 0.740$\pm$0.033 & 0.738$\pm$0.032 & 0.738$\pm$0.036 & 0.792$\pm$0.019 & \bf{0.804$\pm$0.031} \\
\hline
RF & 0.713$\pm$0.097 & 0.788$\pm$0.004 & 0.781$\pm$0.005 & 0.779$\pm$0.006 & 0.78$\pm$0.007 & \bf{0.861$\pm$0.034} & 0.846$\pm$0.035 \\
\hline
MLP & 0.665$\pm$0.110 & 0.724$\pm$0.081 & 0.720$\pm$0.041 & 0.695$\pm$0.065 & 0.714$\pm$0.051 & \bf{0.756$\pm$0.092} & 0.751$\pm$0.078 \\
\hline
LR & 0.703$\pm$0.094 & 0.783$\pm$0.008 & 0.770$\pm$0.009 & 0.775$\pm$0.003 & 0.785$\pm$0.013 & 0.800$\pm$0.012 & \bf{0.802$\pm$0.011} \\
\hline
GNB & 0.377$\pm$0.136 & 0.664$\pm$0.118 & 0.691$\pm$0.009 & \bf{0.694$\pm$0.018} & 0.660$\pm$0.130 & 0.680$\pm$0.123 & 0.666$\pm$0.109 \\
\hline
Tree & 0.688$\pm$0.101 & 0.768$\pm$0.012 & 0.695$\pm$0.016 & 0.699$\pm$0.02 & 0.717$\pm$0.015 & \bf{0.832$\pm$0.023} & 0.827$\pm$0.027 \\
\hline
KNN & 0.689$\pm$0.061 & 0.595$\pm$0.056 & 0.708$\pm$0.035 & 0.688$\pm$0.027 & 0.712$\pm$0.035 & \bf{0.786$\pm$0.038} & 0.783$\pm$0.054\\
\hline
\end{tabular}}
\end{table}
Similarly, imputation improves classification performance in most cases. In large scale imputation, when the missing mechanism is different and more data are absent, mean imputation is no longer as effective as some machine learning methods such as KNN and RF, which gained significant advantages and robustness in our experiments, as highlighted in Table \ref{resulttable}. We also compared Table \ref{resulttable} with Table \ref{whole} and found that cross imputation can outperform single imputation, for example, in averaged results from different classifiers, cross imputation accuracy using RF (76$\%$-80.7$\%$) is significantly higher than that of single imputation(62$\%$-78.1$\%$). This also happens for KNN, mean, and random imputation techniques, indicating that though cross imputation introduces more uncertainty and complexity into analysis, utilisation of full dataset with latent relationships between similar sub-datasets can have a better prediction performance.

\section{Conclusion}
According to our analysis, results from various classifier with single imputation were similar, with only that results from Naïve Bayesian being less precise. Compared with the list-wise deletion, even simple mean imputation can achieve better results. The outcome also suggests that in general RF and MICE are likely to be the best approaches within a small scale database. When the scale enlarges and more uncertainty is introduced, the results indicate that mean imputation is no longer as efficient as machine learning based imputation such as RF and KNN, which are probably the best choices with the least standard deviation. Overall, most of the imputation techniques show strong robustness and high efficiency in cross-dataset imputation with regard to high missingness, outperforming single imputation cases.

\section{Acknowledgement}
This work is fully funded by Melbourne Research Scholarships (MRS), Granted 19/03 and partially supported by Fertility After Cancer Predictor (FoRECAsT) Study. Michelle Peate is currently supported by an MDHS Fellowship, University of Melbourne. The FoRECAsT study is supported by the FoRECAsT consortium and Victorian Government through a Victorian Cancer Agency (Early Career Seed Grant) awarded to Michelle Peate.

%
%
%
%
\bibliographystyle{splncs04}
\bibliography{reference} 

\begin{thebibliography}{10}

\bibitem{Acuna2004missing}
{\sc Acuna, E., and Rodriguez, C.}
\newblock The treatment of missing values and its effect on classifier
  accuracy.
\newblock In {\em Classification, clustering, and data mining applications}.
  Springer, 2004, pp.~639--647.

\bibitem{lung2017}
{\sc Barakat, M.~S., Field, M., Ghose, A., Stirling, D., Holloway, L., Vinod,
  S., Dekker, A., and Thwaites, D.}
\newblock The effect of imputing missing clinical attribute values on training
  lung cancer survival prediction model performance.
\newblock {\em Health information science and systems 5}, 1 (2017), 16.

\bibitem{batista2002study}
{\sc Batista, G.~E., Monard, M.~C., et~al.}
\newblock A study of k-nearest neighbour as an imputation method.
\newblock {\em HIS 87}, 251-260 (2002), 48.

\bibitem{buuren2010mice}
{\sc Buuren, S.~v., and Groothuis-Oudshoorn, K.}
\newblock mice: Multivariate imputation by chained equations in r.
\newblock {\em Journal of statistical software\/} (2010), 1--68.

\bibitem{Moniek2013missing}
{\sc de~Goeij, M.~C., van Diepen, M., Jager, K.~J., Tripepi, G., Zoccali, C.,
  and Dekker, F.~W.}
\newblock Multiple imputation: dealing with missing data.
\newblock {\em Nephrology Dialysis Transplantation 28}, 10 (2013), 2415--2420.

\bibitem{Ives2007pregnancy}
{\sc Ives, A., Saunders, C., Bulsara, M., and Semmens, J.}
\newblock Pregnancy after breast cancer: population based study.
\newblock {\em Bmj 334}, 7586 (2007), 194.

\bibitem{jerez2010missing}
{\sc Jerez, J.~M., Molina, I., Garc{\'\i}a-Laencina, P.~J., Alba, E., Ribelles,
  N., Mart{\'\i}n, M., and Franco, L.}
\newblock Missing data imputation using statistical and machine learning
  methods in a real breast cancer problem.
\newblock {\em Artificial intelligence in medicine 50}, 2 (2010), 105--115.

\bibitem{johnson2006ovarian}
{\sc Johnson, N., Bagrie, E., Coomarasamy, A., Bhattacharya, S., Shelling, A.,
  Jessop, S., Farquhar, C., and Khan, K.}
\newblock Ovarian reserve tests for predicting fertility outcomes for assisted
  reproductive technology: the international systematic collaboration of
  ovarian reserve evaluation protocol for a systematic review of ovarian
  reserve test accuracy.
\newblock {\em BJOG: An International Journal of Obstetrics \& Gynaecology
  113}, 12 (2006), 1472--1480.

\bibitem{Kalton1984random}
{\sc Kalton, G., and Kish, L.}
\newblock Some efficient random imputation methods.
\newblock {\em Communications in Statistics-Theory and Methods 13}, 16 (1984),
  1919--1939.

\bibitem{lee2012transfer}
{\sc Lee, G., Rubinfeld, I., and Syed, Z.}
\newblock Adapting surgical models to individual hospitals using transfer
  learning.
\newblock In {\em 2012 IEEE 12th International Conference on Data Mining
  Workshops\/} (2012), IEEE, pp.~57--63.

\bibitem{lee2009chemotherapy}
{\sc Lee, S., Kil, W.~J., Chun, M., Jung, Y.-S., Kang, S.~Y., Kang, S.-H., and
  Oh, Y.-T.}
\newblock Chemotherapy-related amenorrhea in premenopausalwomen with breast
  cancer.
\newblock {\em Menopause 16}, 1 (2009), 98--103.

\bibitem{Liem2015chemo}
{\sc Liem, G.~S., Mo, F.~K., Pang, E., Suen, J.~J., Tang, N.~L., Lee, K.~M.,
  Yip, C.~H., Tam, W.~H., Ng, R., Koh, J., et~al.}
\newblock Chemotherapy-related amenorrhea and menopause in young chinese breast
  cancer patients: analysis on incidence, risk factors and serum hormone
  profiles.
\newblock {\em PloS one 10}, 10 (2015), e0140842.

\bibitem{lin2019missing}
{\sc Lin, W.-C., and Tsai, C.-F.}
\newblock Missing value imputation: a review and analysis of the literature
  (2006--2017).
\newblock {\em Artificial Intelligence Review\/} (2019), 1--23.

\bibitem{little2019statistical}
{\sc Little, R.~J., and Rubin, D.~B.}
\newblock {\em Statistical analysis with missing data}, vol.~793.
\newblock Wiley, 2019.

\bibitem{Moon1996em}
{\sc Moon, T.~K.}
\newblock The expectation-maximization algorithm.
\newblock {\em IEEE Signal processing magazine 13}, 6 (1996), 47--60.

\bibitem{Nelwamondo2007miss}
{\sc Nelwamondo, F.~V., Mohamed, S., and Marwala, T.}
\newblock Missing data: A comparison of neural network and expectation
  maximization techniques.
\newblock {\em Current Science\/} (2007), 1514--1521.

\bibitem{FoRECAsT}
{\sc Peate, M., and Edib, Z.}
\newblock Fertility after cancer predictor (forecast) study.
\newblock
  \url{https://medicine.unimelb.edu.au/research-groups/obstetrics-and-gynaecology-research/psychosocial-health-wellbeing-research/fertility-after-cancer-predictor-forecast-study}.
\newblock Accessed: 2019-04-15.

\bibitem{peate2011}
{\sc Peate, M., Meiser, B., Friedlander, M., Zorbas, H., Rovelli, S.,
  Sansom-Daly, U., Sangster, J., Hadzi-Pavlovic, D., and Hickey, M.}
\newblock It's now or never: fertility-related knowledge, decision-making
  preferences, and treatment intentions in young women with breast cancer—an
  australian fertility decision aid collaborative group study.
\newblock {\em J Clin Oncol 29}, 13 (2011), 1670--1677.

\bibitem{peate2017fertility}
{\sc Peate, M., Stafford, L., Hickey, M., et~al.}
\newblock Fertility after breast cancer and strategies to help women achieve
  pregnancy.
\newblock In {\em Cancer Forum\/} (2017), vol.~41, The Cancer Council
  Australia, p.~32.

\bibitem{purwar2015hybrid}
{\sc Purwar, A., and Singh, S.~K.}
\newblock Hybrid prediction model with missing value imputation for medical
  data.
\newblock {\em Expert Systems with Applications 42}, 13 (2015), 5621--5631.

\bibitem{rubin2004multiple}
{\sc Rubin, D.~B.}
\newblock {\em Multiple imputation for nonresponse in surveys}, vol.~81.
\newblock John Wiley \& Sons, 2004.

\bibitem{ruddy2014breast}
{\sc Ruddy, K.~J., Gelber, S., Tamimi, R.~M., Schapira, L., Come, S.~E., Meyer,
  M.~E., Winer, E.~P., and Partridge, A.~H.}
\newblock Breast cancer presentation and diagnostic delays in young women.
\newblock {\em Cancer 120}, 1 (2014), 20--25.

\bibitem{schafer1997analysis}
{\sc Schafer, J.~L.}
\newblock {\em Analysis of incomplete multivariate data}.
\newblock Chapman and Hall/CRC, 1997.

\bibitem{stekhoven2012miss}
{\sc Stekhoven, D.~J., and B{\"u}hlmann, P.}
\newblock Missforest—non-parametric missing value imputation for mixed-type
  data.
\newblock {\em Bioinformatics 28}, 1 (2011), 112--118.

\bibitem{van1995python}
{\sc Van~Rossum, G., and Drake~Jr, F.~L.}
\newblock {\em Python tutorial}.
\newblock Centrum voor Wiskunde en Informatica Amsterdam, The Netherlands,
  1995.

\bibitem{wilson1997improved}
{\sc Wilson, D.~R., and Martinez, T.~R.}
\newblock Improved heterogeneous distance functions.
\newblock {\em Journal of artificial intelligence research 6\/} (1997), 1--34.

\end{thebibliography}

\end{document}